\documentclass[conference]{IEEEtran}

\usepackage[colorlinks,urlcolor=blue,linkcolor=blue,citecolor=blue]{hyperref}

\usepackage{color,array}

\usepackage{graphicx}
\UseRawInputEncoding

\usepackage[numbers]{natbib}
\usepackage{multirow}
\usepackage{xcolor}
\usepackage{adjustbox}
\usepackage{dblfloatfix}
\usepackage{graphicx}
\usepackage{tabularx}

\usepackage{tabularray}
\usepackage{pdflscape}
\usepackage{longtable}
\usepackage{colortbl}

\usepackage{array}
\usepackage{amssymb}
\usepackage{amsmath}
\usepackage{mathtools}
\usepackage{comment}
\usepackage{soul}

\begin{document}

\title{Memory-Augmented Graph Neural Networks: A Brain-Inspired Review}

\author{Guixiang Ma \IEEEmembership{Member, IEEE}, Vy A. Vo, Theodore Willke, and Nesreen K. Ahmed \IEEEmembership{Senior Member, IEEE}
\thanks{Manuscript received 26 Jun 2022. \textit{(Guixiang Ma and Vy A. Vo are co-first authors.) (Corresponding authors: Vy A. Vo, Guixiang Ma.)}}
\thanks{G. Ma, V. A. Vo, and T. Willke are with Intel Labs, Intel Corporation, Hillsboro, OR 97124 USA (e-mail: \{guixiang.ma, vy.vo, ted.willke\}@intel.com). }
\thanks{N. K. Ahmed is with Intel Labs, Intel Corporation, Santa Clara, CA 95054 USA (e-mail: nesreen.k.ahmed@intel.com).}
\thanks{This paragraph will include the Associate Editor who handled your paper.}}

\markboth{Journal of IEEE Transactions on Artificial Intelligence, Vol. 00, No. 0, Month 2022}
{Ma and Vo \MakeLowercase{\textit{et al.}}: Memory-Augmented Graph Neural Networks: A Brain-Inspired Review}

\maketitle

\begin{abstract}
We provide a comprehensive review of the existing literature on memory-augmented GNNs. We review these works through the lens of psychology and neuroscience, which has several established theories on how multiple memory systems and mechanisms operate in biological brains. We propose a taxonomy of memory-augmented GNNs and a set of criteria for comparing their memory mechanisms. We also provide critical discussions on the limitations of these works. Finally, we discuss the challenges and future directions for this area.
\end{abstract}

\begin{IEEEImpStatement}
Memory-augmentation of graph neural networks is an emerging research field in the deep graph learning community. These augmentations can enhance GNNs' capabilities for structured representation learning and relational reasoning tasks, such as human-object interaction prediction, question answering, algorithmic reasoning, etc. This paper provides the first systematic review of memory-augmented GNNs from the perspective of neuroscience and psychology. We identify the open challenges in this field and shed light on the promising directions for future work. Our work will help facilitate novel brain-inspired designs to advance graph neural networks for various domains.
\end{IEEEImpStatement}

\begin{IEEEkeywords}
Graph neural networks, Memory-augmented neural networks, Relational reasoning, Structured representation
\end{IEEEkeywords}

\section{Introduction}
\label{sec:intro}

Graph neural networks (GNNs) are a family of models at the intersection of deep learning and structured approaches, and have been widely studied in recent years \cite{wu2020comprehensive, zhou2020graph}. GNNs typically rely on local message passing to learn representations from both irregular graphs (e.g. social networks, molecules) and regular graphs (e.g. image grids). The combination of deep learning and message passing allows them to extract structured information and relational dependencies from graphs to accomplish a variety of tasks \cite{xu2018powerful, battaglia2018relational, wu2020comprehensive, zhou2020graph}. However, a variety of limitations in GNNs have led to a convergent set of approaches that augment a message passing network with some form of memory. Here, we review a diverse set of papers on memory-augmented GNNs 

\begin{figure*}[!b]
    \centering
    \includegraphics[width = 0.75\textwidth]{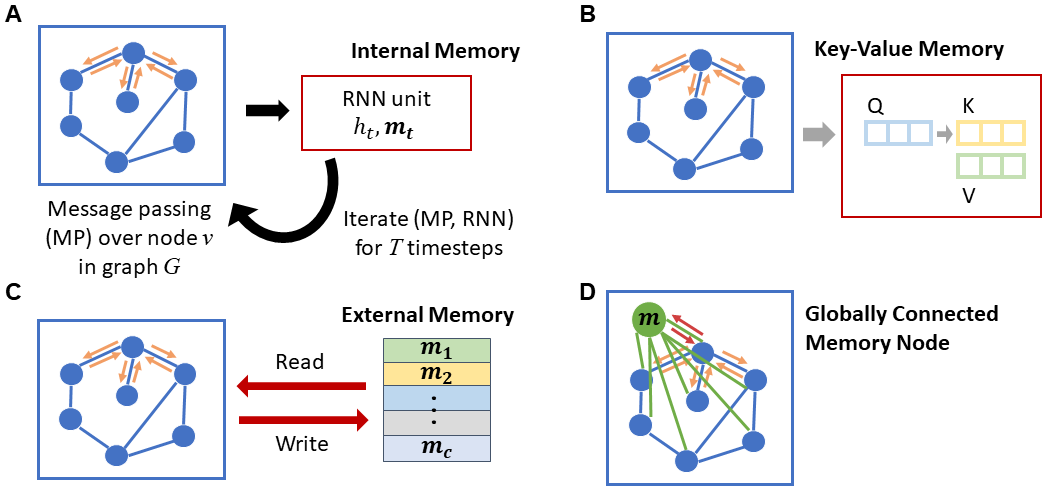}
    \caption{Examples of different forms of memory used with GNNs. \textbf{A, C}] Red boxes outline the network mechanism that contains the memory. \textbf{B, D}] Red arrows indicate how read and write mechanisms interact with the memory state $m$.}
    \label{fig:memoryGNNs}
\end{figure*}

The first motivation to review memory-augmented GNNs from this perspective is historical. Many areas of machine learning have repeatedly drawn inspiration from the brain to build architectures that maintain information over time. Early efforts include recurrent neural network (RNN) architectures \cite{schuster1997bidirectional, hochreiter1997long} that use internal states to store information about sequential inputs. These have made many advances in several areas of machine learning, and are key to several of the memory-augmented GNNs we review here. Other memory-augmented neural networks (MANNs) utilize external memory with differentiable read-write operators that allow the neural network to store and access past experiences. MANNs have been shown to enhance performance on many learning tasks, such as reinforcement learning \cite{pritzel2017neural}, meta learning \cite{santoro2016meta}, and few-shot learning \cite{vinyals2016matching}. The second motivation is that focusing on memory in GNNs may specifically improve performance on structured representation learning problems such as relational reasoning. 
In particular, studies in both cognitive neuroscience and machine learning suggest that memory plays a key role in relational reasoning \cite{whittington2020TolmanEichenbaum}. Several cognitive behavioral studies of human recall have led to proposals that the contents of memory are stored as structured representations \cite{Brady_2011, bradyContextualEffectsVisual2015, naimEmergenceHierarchicalOrganization2019}. These are bolstered by neuroscience findings of a similarity structure in neural representations of memories, i.e. that more similar memories lead to more similar representations
\cite{ezzyatSimilarityBreedsProximity2014, schlichtingMemoryAllocationIntegration2017}. In the realm of brain-inspired machine learning, \cite{santoro2018relational} demonstrated that augmenting an RNN with a relational memory module that allows memories to interact with each other improves its performance for various relational reasoning tasks, including as partially observed reinforcement learning tasks, program evaluation, etc. The final motivation for this review is that GNN researchers have already turned to memory mechanisms to alleviate specific shortcomings of these models. These shortcomings include learning long-range dependencies across large graph neighborhoods, as well as across long-range relationships across time for dynamic graphs. However, these efforts are largely independent and have not been reviewed from a comprehensive or unified perspective. The aim of our review is to create a framework for understanding memory-augmented GNNs that may generate new proposals for network architectures, datasets, tasks, and ML modules that can tackle problems in graph learning and improve the expressiveness of GNNs.

capabilities on various graph learning tasks. We emphasize that this paper does not attempt to review the extensive literature on GNNs or deep graph representation learning that do not explicitly use memory. Prior works have focused on these topics [see \cite{wu2020comprehensive,zhou2020graph,zhang2020deep}]. Here instead, we focus on the body of works that use memory in the model design of graph neural networks and provide a taxonomy to categorize the models. Many of the recent works that we review in this paper have not been discussed in prior surveys of GNNs. We also provide a critical discussion of pros and cons of different model designs to provide practitioners with insights to inform the design of their next memory-augmented GNN. Part of this critical review is undertaken from a neuroscience perspective to provide proposals for brain-inspired memory augmentations. To the best of our knowledge, this is the first review paper on this topic. Our main contributions are summarized below:

Our contributions are:
\begin{itemize}
\item We propose a taxonomy of existing memory-augmented GNNs.
\item We provide a set of criteria for comparing the memory mechanisms, inspired by neuroscience.
\item We present critical discussion on limitations.
\item We provide insights on the open challenges and future directions in this field.
\end{itemize}

\section{Background}
\label{sec:background}

First we will review some key concepts and background of graph neural networks and the historical use of the concept of memory in this literature. Then we will review some key concepts and background on the cognitive neuroscience of memory. Together these two literatures will help establish our taxonomy to understand how memory may improve the performance and expressivity of GNNs.

\subsection{Background: graph neural networks}
Let $G = (V,E)$ denote a graph, where $V$ is the set of nodes, $E \subseteq V \times V$ is the set of edges. Let $N = \mid V\mid$ be the number of nodes, and $\mathbf{A}, \mathbf{D}$ be the adjacency matrix and degree matrix. For each node $v \in V$, let $N_v = \{u\mid(u,v)\in E\} $ be its neighborhood node set, and $\mathbf{x}_v$ be the feature vector of node $v$. While there may also be edge-level and graph-level features in GNNs, we give definitions that omit them for simplicity (see \cite{wu2020comprehensive} for a more detailed definition).

Message passing in GNNs consists of three steps that update the features $X_v$. There is the \emph{message} function $\psi$, the \emph{aggregation} operator $\bigoplus$, and the \emph{readout} function $\phi$. The learnable message function defines how the neighbors' feature vectors are sent to the the target node, $\psi(\mathbf{x}_u, \mathbf{x}_v$. The aggregation operator $\bigoplus$ is a permutation-invariant function such as a sum, mean, or maximum that combines the messages across the neighborhood node set $\mathcal{N}_v$. The learnable readout $\phi$ then updates the hidden node representation. The message and readout functions $\phi$ and $\psi$ can be neural network components like a multilayer perceptron or an attention mechanism \cite{velivckovic2017gat}. The particular definition of these components is often the key distinction between different GNN architectures \cite{xu2018powerful}. These may also be referred to as message-passing neural networks (MPNNs) in the field of geometric deep learning \cite{bronstein2021review}.

\begin{align}
    \forall u \in \mathcal{N}_v, {\mathbf{X}_v = \phi(\mathbf{x}_v,  \bigoplus \psi(\mathbf{x}_v, \mathbf{x}_u)}
\label{eq:COMBINE}
\end{align}

GNNs can be considered a generalization of other popular machine learning approaches, mainly convolution and recurrence \cite{battaglia2018relational, bronstein2021review}. Convolutional networks operate over a grid and are invariant to spatial translation. Recurrent neural networks (RNNs) operate over timesteps and are invariant to time translation. GNNs operate over graphs which represent their nodes and edges as permutation-invariant sets. Both grids and temporal sequences can be represented as graphs, and are hence special cases of GNNs. This has inspired several GNN architectures that use these ML operations as part of their message passing mechanisms, such as early recursive GNNs \cite{scarselli2009gnn} and graph convolutional networks (GCNs) \cite{kipf2016semi}. While their form is general, GNNs must either be equivariant or invariant functions of the data depending on what type of task they are performing \cite{wu2020comprehensive}. This invariance can be an expressive limitation of GNNs and has led to a body of theoretical work to formally define the abilities of these models \cite{murphy2019pooling, maron2018invariant}.

\begin{figure}[b]
    \centering
    \includegraphics[width = 0.95\columnwidth]{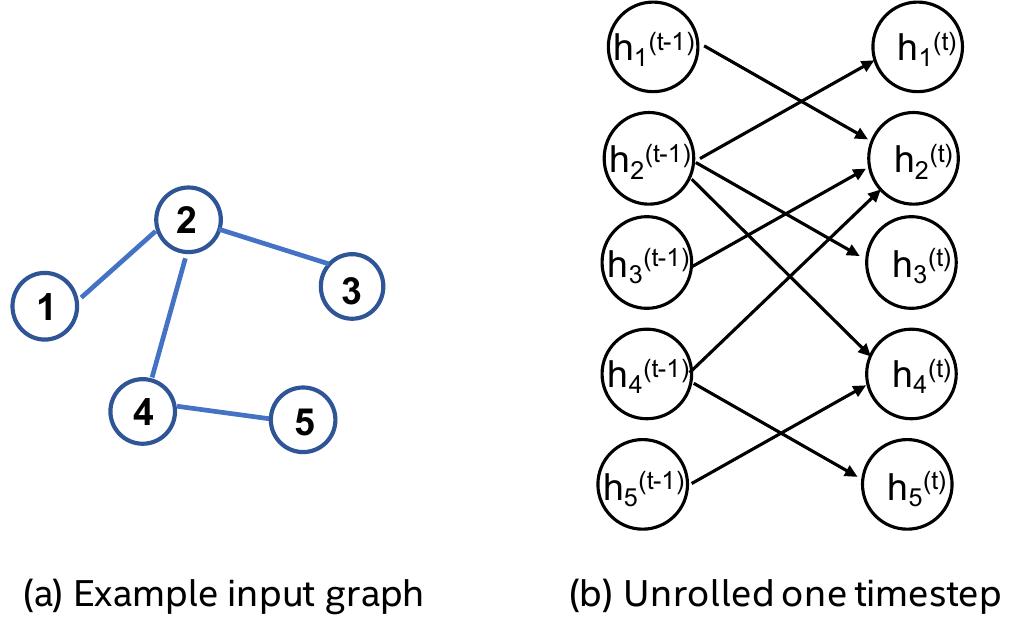}
    \caption{Early memory GNN model. GNNs with recurrent units retain a history of computation in the hidden state. For an input graph shown in (a), we illustrate how information is propagated over one timestep (b).}
    \label{fig:gatedGNN}
\end{figure}

A recent comprehensive review by Wu et al. \cite{wu2020comprehensive} taxonomized GNNs into four main categories. In addition to convolutional and recurrent GNNs, they added spatial-temporal GNNs and graph autoencoders that learn network embeddings. We focus our review on GNNs that have been described as having some form of memory. This includes GNNs in all four of Wu et al.'s categories \cite{wu2020comprehensive}. These memory augmentations might store information about the graph structure or previous timestep information. An illustration of different forms of memory used in augmented GNNs is given in Figure \ref{fig:memoryGNNs}. One early and widely used form of memory in GNNs involved the adaptation of RNN units, such as Gated Recurrent Unit (GRUs), to maintain an internal memory state for graph inputs. In the GatedGNN model, \cite{li2015gated} employ a gated recurrent unit (GRU) \cite{cho2014learning} to maintain a memory for each node. This memory is updated by its previous hidden states and its neighbors' hidden states, which can be defined as:

\begin{align}
    h_v^{(t)} = GRU(h_v^{(t-1)}, \sum_{u\in N(v)} \mathbf{W}h_u^{(t-1)})
\label{eq:GGNN}
\end{align}

where $\mathbf{W}$ represents the learnable weight parameters. An example of the propagation model of gated GNNs for one timestep can be found in Figure \ref{fig:gatedGNN}.

\subsection{Background: memory in psychology and neuroscience}

\begin{figure}[h]
\centering
\includegraphics[width=\columnwidth]{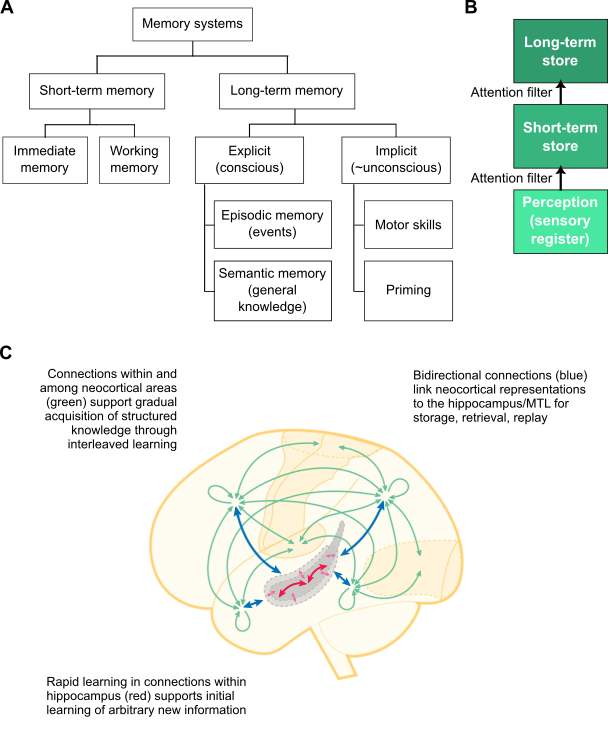}
\caption{Cognitive neuroscience theories on memory systems and their interactions. \textbf{A}] A typical taxonomy of multiple memory systems. \textbf{B}] A classic theory on interaction between perception (sensory register) and memory systems, where information is passed via attentional filters. \textbf{C}] Brain systems involved in learning structured knowledge. Figure adapted from \cite{kumaranEtAl2016CLS}.}
\label{fig:neuro}
\end{figure}

Figure \ref{fig:neuro}A shows a typical taxonomy of multiple memory systems that distinguishes short-term memory from long-term memory \cite{roedigerTypologyMemoryTerms2017}. Long-term memory is required to learn structured knowledge and durable facts about the environment over days to years \cite{Purves_Textbook_2004, Squire_Wixted_2011}. One type of long-term memory, episodic memory, enables us to store details of specific events, or episodes, that may be useful in the future. Both forms of long-term memory are supported by the hippocampus and related medial temporal lobe structures \cite{Squire_2004, Squire_Wixted_2011} (gray structure in Fig. \ref{fig:neuro}C). By contrast, working memory allows us to engage in the short-term manipulation of information, such as adding and multiplying large numbers. Working memory plays a central role in global workspace theory \cite{dehaene1998neuronalGlobalWorkspace, baars2003consciousWM}, which posits the existence of a globally accessible, limited capacity workspace where information can be coordinated and communicated. This global workspace introduces a bottleneck that forces the connected modules to compete, since only the most important or relevant modules at any given time can rewrite the contents of memory in the workspace.

Figure \ref{fig:neuro}B illustrates a classic theory on interaction between perception (sensory register) and memory systems, where attention gates between systems (black arrows) selectively filter what information enters into the next stage \cite{Atkinson_Shiffrin_1968}. These gates allow for attentional selection of the content in memory, so that only relevant information from perception enters short-term memory. Furthermore, only information that would be useful in long-term memory is encoded.
Figure \ref{fig:neuro}C depicts the brain systems thought to be involved in complementary learning systems, which facilitate the transfer of information from short-to long-term memory. The fast learning system is largely thought to rely on synaptic plasticity in the hippocampus (red arrows). The slow system relies on cortical plasticity (green arrows) as well as hippocampal-cortical interactions (blue arrows).
While the theories illustrated in Figure \ref{fig:neuro} build on a molecular and cellular understanding of memory, their main focus is to define cognitive modules or neural systems that tackle a particular set of problems. A similar approach can help shape our understanding and function of memory modules in GNN architectures.

Another key advance was made in understanding the different stages of memory, regardless of its duration. The stages include (1) encoding, (2) maintenance or consolidation, and (3) retrieval \cite{Preston_Wagner_2007, Sweatt_2010}. It is widely understood that some active selection mechanism determines what content is processed in each stage of memory. This selection is necessary because memory capacity is limited -- and is especially necessary in a globally accessible workspace since it has a very low capacity limit compared to long-term memory. Therefore, we can also analyze memory GNN to determine whether there is an active mechanism that selects specific content to write to memory or to retrieve from memory. In some cases, the design of these memory mechanisms defaults to the simplest method, which is to perform no selection at all and allow all content to be written and retrieved. 

\section{Framework and Taxonomy}
\label{sec:framework}

We analyze each memory-augmented GNN by (1) the scope and form of the memory, (2) the machine learning problem it was designed to tackle, and (3) the read, write, and forget mechanisms of the memory (Tables \ref{tab:taxonomy1}-\ref{tab:taxonomy2}).

\label{sec:Taxonomy}

\subsection{Scope of memory}
\label{subsec:memscope}
We described GNNs above as a generalized model form that can maintain invariance over space or time, depending on what is represented by the nodes and edges in the graph. Similarly, memory-augmented GNNs can store information about space, time, or both space and time. In Table \ref{tab:taxonomy1} we define space as distance in the graph.
Memory GNNs which operate over space are storing information about graph structure in memory, e.g. to enable non-local processing or to learn particular structures such as hierarchies. Memory GNNs which operate over time, such as recurrent GNNs, are enabling history-dependent processing.  Memory GNNs that operate over both space and time can process graphs that have evolving features or connectivity over time (e.g., dynamic graphs).

As introduced in Section \ref{sec:background}, Gated GNNs \cite{li2015gated} incorporate a gated recurrent unit and unroll the recurrent computation for some number of timesteps $T$, backpropagating through time to compute the gradients. Unlike other GNNs, they can process sequential information. Therefore we classify them as operating over time, even though they may store spatial or structural information. See Figure \ref{fig:venndiagram} for an illustration of the scope of specific memory GNN models.

The scope of memory is a key distinction to understand when drawing parallels between GNNs and humans. Human memory, by definition, is always operating over the scope of time. Information in human long-term memory, however, can encode relational structures that can enable structural generalization \cite{whittington2020TolmanEichenbaum}.

\subsection{Problems in machine learning}
\label{subsec:mlproblems}
Memory augmentation has been used to address several different types of machine learning problems, which we divide into categories: long-range dependencies, expressivity of message-passing networks, dynamic graphs, and structured representation learning. While a certain work may fall into more than one of these categories, we focus on categorizing the work based on their motivation for memory augmentation.

\subsubsection{Memory for long-range dependencies}
Learning long-range dependencies across a graph is a known problem for message-passing GNNs \citep{alon2020bottleneck, wu2020comprehensive, dwivedi2022lrgb}. This is mainly due to the fact that GNNs update node features based on local information propagated across some $k$-hop neighborhood \cite{xu2018powerful,alon2020bottleneck}. However, many problems in graph learning may require interactions with distant nodes \cite{dwivedi2022lrgb}. Researchers first hypothesized that increasing the depth of the network by stacking $L$ GNN layers would enable learning representation in an $L$-hop neighborhood. However, it has been observed that when the number of layers increases, the node representations become indistinguishable. This is known as the ``over-smoothing" issue \cite{wu2020comprehensive}. Moreover, as the number of layers increases, the number of nodes in each node's receptive field grows exponentially. This also causes ``over-squashing", since information from the exponentially-growing receptive field will be compressed into fixed-length node vectors \cite{alon2020bottleneck}. As a consequence, the graph is not able to propagate messages from distant nodes, and thus only captures short-range information.

Memory augmentation has been proposed to bolster non-local learning abilities in GNNs. One way to do this is to add a virtual node that is connected to every other node in the graph. Gilmer et al. \cite{gilmer2017neural} name this virtual graph element a "master node", and allow each node to both read from and write to the master during every step of message passing. They describe the master node as a "global scratch space" for the GNN -- which we note is a similar idea to global workspace theory in cognitive neuroscience. This virtual node approach is discussed in more detail in the next section on expressivity.

In addition to memory augmentation that operates over the scope of space, other works have added memory to GNNs to enable learning long-range dependencies over time. For example, Rossi et al. \cite{rossi2020temporal} create a memory state that stores a vector for each node the model has encountered so far. This memory is not global -- that is, it does not store information about the entire graph. Another work proposes a GNN architecture for video processing, where the memory state stores previous information about some portion of the video frame \cite{nicolicioiu2019recurrent}.

\subsubsection{Memory for expressivity}
Theoretical work on the expressivity of GNNs often explores their ability to distinguish different graph structures. \cite{xu2018powerful} showed that message-passing GNNs can be as powerful as the Weisfeiler-Lehman (WL) graph isomorphism test, which is an algorithm to distinguish whether two graphs are topologically identical. Like GNNs, the WL test also aggregates features from neighboring nodes, thereby mapping different node neighborhoods to different feature vectors. However, this analysis assumes that the node features are countable. See \cite{corso2020pna} for a discussion on extending work on GNN expressiveness to continuous feature spaces.

In memory-augmented GNNs, the only theoretical work we found explored the impact of globally connected (GC) virtual nodes on expressiveness. These GC nodes, introduced by Gilmer et al. \cite{gilmer2017neural}, are connected to every other node in the graph and are updated via message passing. Xiong et al. \cite{xiong2020memory} explored how adding one or more of these GC nodes can act as a memory to improve performance on node classification tasks. Due to the permutation invariance of message passing, GNNs can fail to distinguish graph substructures that are locally isomorphic -- that is, locally isomorphic neighborhoods will map to identical feature vectors. They show that defining GC memory nodes can break the symmetry in locally indistinguishable graph substructures. This is accomplished by defining new aggregation rules:

\begin{alignat}{3}
    \forall u \in \mathcal{N}_v, \forall i \in \mathcal{V}_\mathcal{M}&, {}& &{\mathbf{h}_v} &= &\bigoplus(\mathbf{x}_v, \mathbf{x}_u, \mathbf{m}_i) \\
                                 \forall i \in \mathcal{V}_\mathcal{M}&, {}& &{\mathbf{m}_i} &= &\bigoplus(\mathbf{x}_v, \mathbf{m}_i)
\label{eq:virtualnode}
\end{alignat}

where $\mathcal{V}_\mathcal{M}$ is the set of memory nodes, and $\mathbf{m}_i$ is the representation for memory node $i$. Xiong et al. \cite{xiong2020memory} define two possible models with memory nodes, and further show that this improved performance on standard benchmarks.

Recently \cite{hwang2022VNlinkpred} investigated a variant on this concept by augmenting networks with multiple virtual nodes. Rather than being globally connected, each virtual node is connected to a different group of nodes. They also show that these virtual node augmentations enable the GNN to distinguish between isomorphic graph structures. Furthermore, their analysis suggests that using multiple virtual nodes instead of a single GC node can have a greater effect on learning long-range dependencies and alleviating over-smoothing \cite{hwang2022VNlinkpred}.

Cai et al. \cite{cai2023connectionMPNNTransformer} establish a link between message-passing networks with GC nodes and two other related deep learning approaches: self-attention transformer layers and DeepSets \cite{zaheer2017deepsets}. We dive deeper into this connection in Section \ref{sec:otherrelated}.

\subsubsection{Memory for dynamic graphs}
Dynamic or temporal graphs allow the nodes, features, and edges to change over time. These often depend on sequence models with recurrence and internal memory representation \cite{gaoribeiro2022}. A review of all the work on temporal graph learning that employs RNNs is out of the scope of this paper. Instead we describe two general approaches to temporal graphs, following Gao and Ribiero \cite{gaoribeiro2022}. The typical approach treats a dynamic graph as a series of static graph snapshots, and applies a GNN to each snapshot. Then an RNN cell might operate over these outputs. The memory state of the RNN cell can be understood to encode the previous graph snapshot. In contrast, other methods first apply sequence models to the node and/or edge features, then apply a GNN to the hidden representations of the sequence model \cite{gaoribeiro2022, xu2020inductive, rossi2020temporal}. Here the memory states of the RNN encode the history of the feature representations, and the GNN aggregates over this history. This is the case for the Temporal Graph Network \cite{rossi2020temporal}, which uses this memory to learn long-range temporal dependencies.

One can also view the evolving graph as a persistent data structure. Following this line of thought, \cite{strathmann2021persistent} propose an approach that allows a GNN to learn which nodes and features should persist to complete an algorithmic reasoning task. They suggest this is akin to episodic long-term memory.

\subsubsection{Memory that captures relational structure}
The last category of approaches views the graph itself as memory, and devises an augmented GNN to learn the relational structure. One work proposes memory layers in their GNN that can learn hierarchical representations \cite{khasahmadi2020memory}. They use the term memory in analogy to memory-augmented neural networks that formulate a series of matrix products as query-key-value memory, wherein a query representation $\mathbf{Q}$ is used to find memory keys $\mathbf{K}$. These are used to retrieve the actual values for the information in memory, $\mathbf{V}$.

Other work represents an image or spatial environment as a graph, and stores this graph in memory to perform tasks like visual question answering \cite{khademi2020multimodal} or spatial navigation \cite{zweig2020neural}. Another approach in this category considers the information in a knowledge graph to be stored in memory \cite{moon2019memory, wu2022GraphMemDialog}.

\RenewDocumentCommand\TblrAlignLeft{}{\RaggedRight}

\definecolor{Silver}{rgb}{0.752,0.752,0.752}
\begin{table*}[!ht]
\caption{Form and scope of memory GNNs for different problems and applications.}
\begin{tblr}{
  width = \textwidth,
  colspec = {Q[100]Q[50]Q[150]Q[125]Q[75]},
  row{1} = {Silver, c},
  hlines,
}
\textbf{Model} & \textbf{Scope} & \textbf{ML problem addressed by memory} & \textbf{Form of Memory} & \textbf{Application}\\
Graph Memory Network (GMN), MemGNN \cite{khasahmadi2020memory} & space & Capturing relational structure (hierarchy) & key-value store & Graph classification\\
MemGAT, MemGCN \cite{xiong2020memory} & space & Expressivity (local isomorphism) & global information store & Node classification\\
Multimodal Neural-GMN \cite{khademi2020multimodal} & space & Capturing relational structure & spatially organized external memory \& internal memory & Visual QA\\
Memory GN (MGN) \cite{moon2019memory} & space & Capturing relational structure (KGs) & knowledge graph store & Question Answering\\
Memory-augmented GNN (MA-GNN) \cite{ma2020memory} & time & Long-range dependencies & cache (short-term memory, STM) and key-value store (long-term memory, LTM) & Sequential recommendation\\
Gated GNN \cite{li2015gated} & time & Long-range dependencies; Sequential outputs & internal & Program verification\\
TreeLSTM \cite{tai2015treelstm} & time & Capturing relational structure (hierarchy); long-range dependencies & internal & Sentence parsing\\
CacheGNN \cite{ma2022cachegnn} & time & Long-range dependencies & internal cache & Node classification\\
Recurrent Space-Time GN (RSTG) \cite{nicolicioiu2019recurrent} & space-time & Dynamic graphs; long-range dependencies & internal & Video understanding\\
Persistent Message Passing (PMP) \cite{strathmann2021persistent} & space-time & Dynamic graphs & internal & Algorithm reasoning\\
Temporal GN (TGN) \cite{rossi2020temporal} & space-time & Dynamic graphs; long-range dependencies & internal & Dynamic node classification, Link prediction\\
GraphMem-Dialog \cite{wu2022GraphMemDialog} & space-time & Capturing relational structure (KGs); long-range dependencies & internal \& external & Dialog systems\\
Zweig et al. \cite{zweig2020neural} & space-time & Capturing relational structure & internal to the graph & Graph navigation\\
GC master node\cite{gilmer2017neural} & space & Long-range dependencies; expressivity & global information store & Quantum chemistry\\
Virtual supernodes \cite{hwang2022VNlinkpred} & space & Expressivity & external & Link prediction \\
\end{tblr}
\label{tab:taxonomy1}
\end{table*}

\begin{table*}[!h]
\caption{Mechanisms for storing and modifying information in memory GNNs.}
\begin{tblr}{
  width = \textwidth,
  colspec = {Q[100]Q[150]Q[75]Q[175]},
  row{1} = {Silver,c},
  hlines,
}
\textbf{Model} & \textbf{Write Mechanism} & \textbf{Forgetting Behavior} & \textbf{Retrieval Mechanism}\\
Graph Memory Network (GMN), MemGNN \cite{khasahmadi2020memory} & cluster assignment & N/A & similarity between query and memory nodes\\
MemGAT, MemGCN \cite{xiong2020memory} & depends on aggregation rule & N/A & all nodes read memory from previous layer\\
Multimodal Neural-GMN \cite{khademi2020multimodal} & aggregates multimodal information preprocessed through GRUs & reset or forget gate & all nodes read memory from previous timestep\\
Memory GN (MGN) \cite{moon2019memory} & None & None & query-memory similarity, then graph walk\\
Memory-augmented GNN (MA-GNN) \cite{ma2020memory} & STM: store most recent items; LTM: train key-value embedding & STM: remove oldest items; LTM: none specified & STM: entire cache used; LTM: attention embeds query\\
Gated GNN \cite{li2015gated} & learnable weights on previous time-step information & reset or forget gate & all nodes read memory from previous timestep\\
TreeLSTM \cite{tai2015treelstm} & learnable weights on neighbors' information & forget gate & all nodes read memory from neighbors\\
CacheGNN \cite{ma2022cachegnn} & stores most recent hidden states & remove oldest & use all stored states\\
Recurrent Space-Time GN (RSTG) \cite{nicolicioiu2019recurrent} & learnable weights on previous time-step information & forget gate & all nodes read memory from previous timestep\\
Persistent Message Passing (PMP) \cite{strathmann2021persistent} & learnable persistency mask on previous hidden states & None & learnable relevance mask\\
Temporal GN (TGN) \cite{rossi2020temporal} & message function; message aggregation over time; learnable memory update function   (GRU) & reset or forget gate; temporal embedding module & temporal graph attention/sum\\
GraphMem-Dialog \cite{wu2022GraphMemDialog} & learnable weights on previous time-step and current neighbors' information; write controller for external memory & update gate & a learnable read controller on previous time-step memory cells; attention on entities in memory\\
Zweig et al. \cite{zweig2020neural} & learnable weights on previous time-step information & reset or forget gate & all nodes read memory from previous timestep\\
GC master node \cite{gilmer2017neural} & learnable memory update function (GRU) & reset gate & variable length processing on memory embeddings with an attention step\\
Virtual supernodes \cite{hwang2022VNlinkpred} & cluster assignment, MP definition & depends on MP definitions & all neighbors (determined by cluster assignment)
\end{tblr}
\label{tab:taxonomy2}
\end{table*}

\begin{figure}[t!]
\centering
\includegraphics[width=\linewidth]{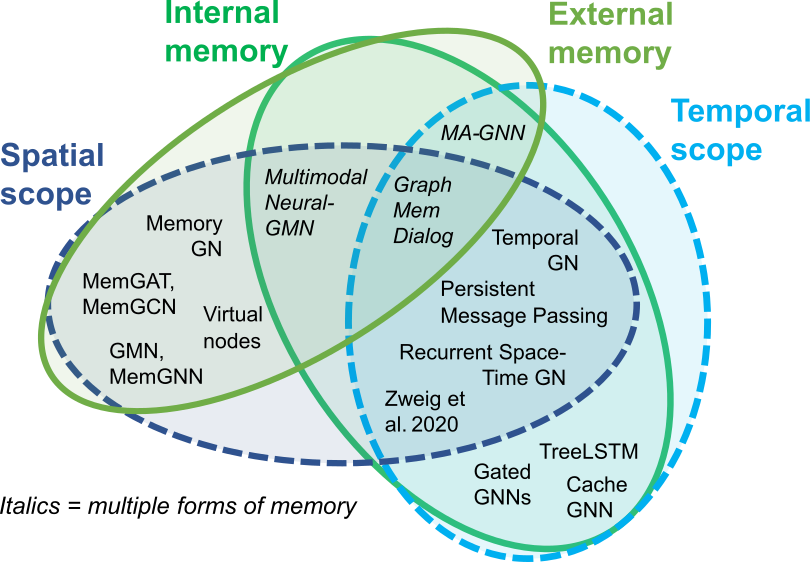}
\caption{Scope and form of memory GNNs reviewed here. All of the works shown in Table \ref{tab:taxonomy1} are depicted here, sorted by the scope of the memory (Section \ref{subsec:memscope}) and whether it is internal or external (Section \ref{subsec:memtypes}).}
\label{fig:venndiagram}
\end{figure}
\subsection{Forms of memory: internal, external, and more}
\label{subsec:memtypes}

By analogy to taxonomies of human memory, we describe a few different forms of memory (Figures \ref{fig:memoryGNNs} and \ref{fig:venndiagram}).

\subsubsection{Internal memory} There is a sizeable body of work on GNNs which rely on a recurrent unit to process some input for a fixed number of timesteps $T$. The recurrent unit contains a hidden state, which is a short-term memory store that propagates information from timestep to timestep based on a rule such as the one defined by Equation (\ref{eq:GGNN}) and illustrated in Figure \ref{fig:gatedGNN}. In a typical gated GNN, hidden states from $t < -T$ are discarded. Given that these hidden states are temporary memories directly mapped to a single node, we can view them as memories that are \textit{internal} to the nodes. A recent work which caches these node-specific hidden states are also only storing internal memories \cite{ma2022cachegnn}.
By analogy to biological systems, an internal memory state can be thought of as persistent activity in a single neuron, whereas an external memory state would be a record of activity across a population of neurons. From this we can see that any information that would be beneficial to many nodes should be stored in external memory. On the other hand, it would be inefficient to store information that is only useful to a single node or a small number of nodes in external memory -- instead, storing it in a state internal to the node is preferable. The term external memory is in reference to MANNs such as the Neural Turing Machine \cite{graves2014neural}, where a controller learns what to read and write to memory using inputs across many neurons. The external memory has a defined size, or capacity, that sets how many discrete memory vectors are stored. In an augmented GNN for dialog systems, for example, the controller uses input from recent context and from a knowledge base to figure out what to read and write to the memory state \cite{wu2022GraphMemDialog}. 
\paragraph{Other forms of external memory} As described in Section \ref{subsec:mlproblems}, one augmentation to the graph structure that can be considered a form of globally accessible memory is the virtual node which is connected to every single other node in the graph. Unlike the external memories modeled after MANNs, the read and write rules for GC memory nodes are defined by the message passing paradigm. Since these memories span many units, we classify them as external memories in Figure \ref{fig:venndiagram}. However, their distinct form warrants a separate entry in Table \ref{tab:taxonomy1}.

A few memory-augmented GNN works reference the query-key-value memory framework, which has been widely used in other deep learning networks \cite{miller2016KVmemnetworks}. Here, the memory can be organized by the structure between the key vectors, each of which is tied to a value vector that represents a retrieved memory. Typically the retrieval mechanism relies on a weighted combination across value vectors. These representations are not tied to specific nodes in the graph. We adopt the same convention as above -- while these are external memories by our definition, we label them as distinct forms in Table \ref{tab:taxonomy1}.

Similar, work that treats a knowledge graph as a memory store is necessarily storing some information across nodes, and cannot be considered internal memory.

\subsection{Memory Write Mechanism}
All memory modules require a write mechanism, which determines what information is stored or encoded in memory.
In most cases, the neural network architecture has a rule which determines what is stored in memory. In regimes where memory capacity is limited, a selective write mechanism ensures that only relevant and useful information is stored for later processing. This is in line with cognitive neuroscience accounts of how and why attention interacts with human memory encoding \cite{chunturkbrowne2007}. We found four principal ways that memory-augmented GNNs select information to be written to memory (Table \ref{tab:taxonomy2}. This excludes work which considers the entire graph to be a memory store, and never modifies it \cite{moon2019memory} -- this is considered to have no separate write mechanism. The four methods are modeled after different ML frameworks: caching, gated recurrence, key-value memories, and message passing. We analyze when each method may be useful to guide ML practitioners in the design choices of their next memory-augmented GNN.

\subsubsection{Caching} This is a simple rule for selecting information to store in memory: only the most recent $t$ timesteps are stored \cite{ma2020memory, ma2022cachegnn}. This is an effective choice if it is known that the task would not benefit from storing long-term information, or if the rule is only used for a short-term memory module that will later be combined with a long-term memory module.

For GNNs with internal memory, most used a write rule determined by the gate equations of recurrent units. Here the information written to memory depended on learnable weights in a gated recurrent unit that combine the input with the current hidden state \cite{li2015gated, nicolicioiu2019recurrent, zweig2020neural, wu2022GraphMemDialog}, allowing context-dependent memory encoding.

\subsubsection{Key-value memory} One common form of external memory in neural networks is a key-value store, where the items in memory are stored as a key $k_i$ associated with a value $v_i$ \cite{miller2016kvmemory, zhang2017dynamickvm, huang2018kvmem}. These items are then retrieved based on some input query $q$. The write mechanism can change the entire key-value pair.

The write mechanism consisted of arranging the keys as clusters of the queries, and defining the associated values as projections of the keys back into the query space. Clustering mechanisms are useful when it is known that the information to be stored in memory has some higher-level structure. 

Unlike \cite{khasahmadi2020memory}, \cite{ma2020memory} defined a key-value memory over time to learn a long-term embedding representation of input items. Inspired by previous work on memory \cite{sukhbaatar2015, zhang2017dynamickvm}, they simply train the key-value matrices $K$ and $V$ through backpropagation to write to the memory. While this write mechanism is quite general and does not make assumptions about the structure of memory, it does assume one of the following: (1) once trained, the embedding is an effective static representation of the input data, or (2) the embedding is retrained or fine-tuned whenever the input data statistics change. This type of write mechanism would not be effective in online learning settings.

\subsubsection{Message passing}
A few papers incorporate some component of message passing to update the contents of memory. For example, work that adds a virtual memory node $\mathbf{m}_i$ \cite{xiong2020memory} relies on message passing and aggregation operators to update the hidden state of the virtual node. This method does not directly incorporate graph structure since the virtual node is fully connected. Its main advantage may be the ease of implementation.

\cite{strathmann2021persistent} performed a round of message passing to compute candidate latent features for the next timestep, $\hat{H}^{(t)}$. These served as inputs to predict a persistency mask, which determined which latent states were kept in memory. Again this allows the memory to incorporate the graph structure.
Interestingly, the persistency mask also depended on the retrieval mechanism (see Section \ref{subsec:retrieval}).

In \cite{hwang2022VNlinkpred}, the authors add $s$ virtual nodes by using a graph clustering algorithm to define $s$ clusters of nodes, each of which is fully connected to its own virtual node. Then an aggregation rule determines how information from each node in a cluster is aggregated and combined to update the corresponding virtual node.

\subsection{Memory Retrieval Mechanism}
\label{subsec:retrieval}
in human memory systems there is typically a selective retrieval mechanism that limits irrelevant information from being used in computation \cite{Atkinson_Shiffrin_1968, Ciaramelli_Grady_Moscovitch_2008}. We find that selective retrieval mechanisms in memory-augmented GNNs typically depend on similarity computations, although we discuss two other instantiations here as well. This relative lack of diversity in retrieval mechanisms suggests that practitioners may benefit from considering more flexible methods of memory retrieval, a point we discuss further in Section \ref{sec:future}.

\subsubsection{Similarity-based retrieval} One common retrieval rule depends on the similarity between an input query and the information in memory \cite{khasahmadi2020memory, moon2019memory, ma2020memory, wu2022GraphMemDialog}. This assumes that highly similar information is likely to be relevant to the task. In the MANN literature, this is often referred to as "content-based addressing" \cite{graves2016dnc}. As a result of its relation to MANNs, this retrieval mechanism is often used in systems with memory controllers such as \cite{wu2022GraphMemDialog}. Similarity-based retrieval is also the typical mechanism used for key-value memory modules. While effective, content similarity is only one way of determining what information in memory is relevant. As noted in other memory work outside of the GNN literature, biological evidence suggests that memories encoded around the same time may be useful to retrieve together, even when they are not similar in content \cite{graves2016dnc}. This may be particularly useful to consider in dynamic graph settings, where distant parts of the graph that appear at the same time should be retrieved together.

\subsubsection{Other retrieval mechanisms} In \cite{strathmann2021persistent}, a linear network is used to determine which node-specific features are relevant for retrieval. Unlike other work we reviewed, the authors used a setting where there are ground-truth relevance labels, allowing them to simply train the linear retrieval network with a cross-entropy loss. This is a more generalized approach than a similarity operation, which has its own inductive bias for content similarity. Interestingly, the retrieval mechanism interacts with write mechanism: only relevant nodes are written to memory.

Some memory GNNs have multiple steps as part of the retrieval mechanism. These multi-step mechanisms can exploit the various advantages of different retrieval mechanisms when they are needed. In \cite{ma2020memory}, for example, the query to the long-term memory module was computed by a position-dependent attention mechanism, based on the equations used in Transformer networks \cite{vaswani2017transformers}. Then the similarity between the query and key-value memory was computed. Finally, this long-term memory is aggregated with all the information in the short-term memory cache using a learnable, context-dependent gate. \cite{rossi2020temporal} actually propose several different retrieval mechanisms which range from a simple identity function (read from all the memories, without transforming them) to a a multi-layer graph attention network which aggregates over a temporal neighborhood. Other than the identity function, the suggested retrieval mechanisms in \cite{rossi2020temporal} all take advantage of some time encoding so that the temporal neighborhood of the target node is taken into account.

\subsection{Memory Capacity}
In memory-augmented neural networks, another key factor that may determine the performance or expressiveness of models is their memory capacity, i.e. how much information they can store. For RNNs, for example, two questions about their memory capacity are: (1) How much information about the task can they store in their parameters? (2) How much information about the input history can they store in their units? \cite{collins2016capacity}. The answers to these two questions usually also indicate the potential bottlenecks of these models.

For some memory-augmented GNN models, discrete items or states are stored in their memory -- therefore the capacity is bounded by the number of memory nodes and hidden dimension. For instance, the memory capacity of the model in \cite{xiong2020memory} is bounded by $M \times D$, where $M$ is the number of memory nodes added to the graph, and $D$ is the size of the hidden dimension for the graph neural network layers. In \cite{khademi2020multimodal}, the memory capacity is bounded by $P \times Q \times D$, where $P \times Q$ is the number of regions of the 2D image, and $D$ is the size of the hidden dimension. In \cite{khasahmadi2020memory}, the memory capacity of the proposed memory-based GNN is $M \times D$, where $M$ is the number of centroid nodes in the memory layer of the model, and $D$ is the size of the hidden dimension. The memory graph networks applied in \cite{moon2019memory} store a knowledge graph in its external memory, therefore its memory capacity is the size of the knowledge graph, i.e., $G(V,E)$, where $V$ is the set of nodes that represent entities (e.g. locations, events, public entities) and $E$ is the set of edges that represent the connections among the entities. Similarly, in \cite{wu2022GraphMemDialog}, the external memory is organized as a knowledge graph $G(V,E)$, where nodes in $V$ represent entities and edges in $E$ represent relations between entities. The memory capacity in this case is also the size of the graph $G(V,E)$.    

For the models with recurrent memory, they can store the latent representation of nodes for a number of time steps. As shown in Table I, some of these memory GNNs operate over time, like the models proposed in \cite{ma2020memory,li2015gated} and some others operate over both space and time \cite{nicolicioiu2019recurrent,strathmann2021persistent,rossi2020temporal}. The memory capacity of these models depends on the number of recurrent units, number of time steps, and the number of layers. These also determine how well the models can learn over space or time. However, empirical work suggests that RNN capacity seems to be  limited by the trainability of the architecture, rather than the number of parameters or hidden units \cite{collins2016capacity}.

\subsection{Memory Forgetting}
Memory systems with a limited capacity need to have some mechanism or policy in place to discard information that is no longer useful or relevant. The forgetting might be passive, e.g. memory traces that decay slowly over time, or active, wherein specific information is marked as no longer relevant \cite{richards2017persistence}. In contrast to persistence, forgetting is information transience. Neurobiological studies of memory also indicate that transient memory is important for memory-guided decision-making in dynamic, noisy environments, because transience enhances flexibility by reducing the influence of outdated information and preventing overfitting to specific past events, thus promoting generalization \cite{richards2017persistence}. Our review finds that dedicated forgetting mechanisms, other than the gates in recurrent units, are not common in memory-augmented GNNs. They may be especially important to consider in dynamic graph settings.

Memory GNNs that rely on gated recurrent units have a built-in reset or forget gate to control what information is discarded \cite{khademi2020multimodal,li2015gated,rossi2020temporal,zweig2020neural,nicolicioiu2019recurrent}. \cite{wu2022GraphMemDialog} relies on a memory controller rather than a recurrent unit, but there is no forgetting mechanism separate from the rules used to write or update the memory. It is worth noting that while the persistent message passing model proposed in \cite{strathmann2021persistent} also has internal memory, it does not have any forgetting behavior. Rather, it maintains a persistent set of internal states to represent the history of operations for entities through their lifetime. This kind of behavior can lead to intractable storage costs and complexity issues. Moreover, without a forgetting mechanism, the model can fail when the information stored in memory becomes outdated and not relevant to the current state. In \cite{ma2020memory}, the oldest items stored in the short-term memory are removed and replaced by recent items, while the long-term memory does not have any specified forgetting behavior.

The influence of outdated information is known as the staleness issue in dynamic graphs \cite{kazemi2020representation}. When a node no longer has events which influence or update its internal state, that state may no longer be useful. Rossi et al. \cite{rossi2020temporal} propose a temporal embedding module to mitigate this issue, and suggest different instantiations of this module. All of the methods involving some time encoding that can keep track of the elapsed time since the last event. The GNN-based methods use this time encoding to aggregate information from the memory of neighboring nodes when too much time has elapsed. This allows the network to update (i.e. forget) old information.

We found that forgetting is not useful for memory GNNs that use memory for storing spatial information about a graph, such as the key-value store of graph hierarchical structure in \cite{khasahmadi2020memory} and the global information store in \cite{xiong2020memory}. Here the spatial features they store involve the entire graph, and all that information is required by the models during the feature propagation over the graph. They are also not applicable to memory GNNs that do not have write mechanisms. In \cite{moon2019memory}, the memory was a static knowledge graph that stores entities and their relationships from the episodes in a dataset collection. However, as discussed above, the memory was never modified by the GNN, and it does not have any mechanism for forgetting either.

\subsection{Productive research directions suggested by taxonomy}

Our review and categorization of memory-augmented GNNs using the taxonomic criteria in Tables \ref{tab:taxonomy1} and \ref{tab:taxonomy2} suggest several areas of research that would benefit from further investigation.

\subsubsection{Theoretical Analysis}

Most memory-augmented GNNs demonstrate success through experimental evaluations for specific applications. Few provide theoretical proof or analysis on the impact of memory-augmentation for graph learning tasks. For example, augmenting the GNN with virtual nodes has been empirically investigated in many works \cite{gilmer2017neural, hu2021ogb}. While there has been some theoretical analysis on this topic \cite{xiong2020memory, hwang2022VNlinkpred}, many open questions remain. For example, it would be useful to see theoretical analysis on how virtual memory nodes affect the receptive field \cite{micheli2009neural} of nodes to prove the impact of this memory augmentation for long-range problems.

In other areas of machine learning, there has been analysis on the relationship between self-attention and various forms of memory \cite{brickenPehlevan2021attentionsdm, ramsauer2020hopfield}. Some work examines recurrent models with internal \cite{levy2018lstmdynamic} or external memory \cite{merrill2020hierarchyRNN}. We argue that extending the theoretical analysis of attention and memory to message-passing GNNs will be fruitful. Some work on this has already begun, with a recent paper investigating the relationship between MPNNs with GC virtual nodes and self-attention transformer layers \cite{cai2023connectionMPNNTransformer}, particularly through a connection with the DeepSets \cite{zaheer2017deepsets} model. They find that a MPNN with a GC virtual node can approximate a graph transformer layer.

Beyond attention and memory, however, there are plenty of other questions to answer about the theoretical and practical abilities of GNNs. For example, the impact of gated internal memory on the expressivity of message passing networks is not well-understood. Empirically, \cite{alon2020bottleneck} show that a GatedGNN performs relatively well on a synthetic task that requires learning long-range dependencies, partially addressing the problem of oversquashing. Yet is unknown how this happens -- does adding internal memory change the Ricci curvature of the graph, which is known to affect oversquashing \cite{topping2022riccicurvature}? Or is there another mechanism by which this occurs?

\subsubsection{Graph-structured memories}

The major forms of memory we identified were internal or external to the graph nodes. These memory representations could be modified with writing and forgetting mechanisms. Occasionally the memory was organized in a particular way based on a predefined inductive bias, e.g. a grid representation of an image \cite{khademi2020multimodal}, or clustering similar nodes together \cite{khasahmadi2020memory, hwang2022VNlinkpred}. Alternatively, a knowledge graph was considered itself to be the memory and served as the input to a GNN -- but the knowledge graph itself could not be modified \cite{moon2019memory, wu2022GraphMemDialog}. These two major setups are at odds with the neuroscience literature, which suggests that memories are stored with relational structure that can be modified over time \cite{Brady_2011, bradyContextualEffectsVisual2015, naimEmergenceHierarchicalOrganization2019, ezzyatSimilarityBreedsProximity2014, schlichtingMemoryAllocationIntegration2017, whittington2020TolmanEichenbaum}. Future work should consider building a memory module that is a dynamic, modifiable graph that encodes relational structure. Writing to this memory could involve modifying nodes, edges, and features. In contrast to the standard dynamic graph setting, where the input data evolve over time, the graph-structured memory would represent an internal world model for the GNN that evolves over time and can be updated as new inputs arrive. We note that other MANNs have devised ways to encode relationships between memories -- but these rely on self-attention, rather than an explicit graph structure, to achieve relational structure in memory \cite{santoro2018relational,le2020SAM}.

\subsubsection{Memory mechanisms for temporal graphs}

Temporal graphs pose their own unique set of issues that can be addressed by more human-like memory mechanisms. For example, having a high-capacity long-term memory is likely useful when processing a dynamic graph, since past information could suddenly become relevant. Rather than rely on a selective write mechanism to filter out irrelevant information, then, one should consider more selective retrieval mechanisms. In addition to retrieving information based on content similarity, one could retrieve based on other factors like contextual similarity or temporal encoding similarity. These factors have shown to be key in neuroscience research on long-term memory \cite{Howard_Eichenbaum_2013}.

Furthermore, we proposed above that the memory itself could be a dynamic graph. This would be useful for a setting where inputs evolve over time, and an internal model of the world needs to be continuously updated. The retrieval mechanism could itself be a GNN.

\subsubsection{Multiple memory systems} 

Despite the advances of the memory graph neural network models, there is still a gap between what current models can achieve and human intelligence. The challenges are especially apparent in tasks that require complex language and scene understanding, reasoning about structured data, and transferring learning beyond the training conditions \cite{battaglia2018relational,garg2020generalization}. For instance, we humans have remarkable abilities to remember information over long time, and we can make use of memories at varying time-scales, but current memory neural network models, such as the memory GNN models based on RNNs that only have internal memory \cite{li2015gated,nicolicioiu2019recurrent,strathmann2021persistent,rossi2020temporal}, are still limited in their capabilities to capture long time-scale information. 

One promising direction to further advance memory GNNs in these aspects is to design models with multiple memory systems. As discussed in previous sections, different forms of memory (e.g. internal memory, external memory) have their own properties and advantages, and they can improve different aspects of graph learning tasks. Having a memory GNN with multiple forms of memory could combine their benefits for improving GNN's expressive power, and would mimic the multiple memory systems that exist in biological systems. A similar point on multiple memory systems has been made for models of language processing \cite{nematzadehMemoryHumanArtificial2020}. 
Some of the existing memory GNNs we reviewed have multiple forms of memory, such as a spatially organized external memory and internal GRU memory in the Multimodal Neural-GMN \cite{khademi2020multimodal}, or the external knowledge base combined with a GRU response decoder in the GraphMemDialog model \cite{wu2022GraphMemDialog}. However, these networks are designed specifically for those applications, and there is still a lack of general multiple memory models for GNNs to improve their expressive power to address fundamental problems in graph learning.

According to dominant theories of human learning and memory, there are interactions between different memory systems, as illustrated in Figure~\ref{fig:neuro}B-\ref{fig:neuro}C. For instance, rapidly learned information can be consolidated as structured, long-term memories through bidirectional interactions between the hippocampus and cortex \cite{kumaran2016learning}. When designing multiple-memory-augmented GNN models, it will be key to consider what forms of memories to use and how to formulate interactions between these memories to improve graph learning problems.

\section{Applications}
\label{sec:application}

Different forms of memory or mechanisms may be best suited to a particular type of application. The existing memory GNN works study a variety of tasks and applications (Table \ref{tab:taxonomy1}), which we describe in greater detail here. We also highlight two other graph learning areas that could benefit from memory augmentation: meta-learning and reinforcement learning.

These applications use memory over a spatial scope and may not need forgetting mechanisms (depending on whether the graph is static or dynamic). However, it may still be the case that selective writing and retrieval mechanisms that overweight certain features may improve task performance.

\paragraph{Node Classification.} Node classification is the task of predicting class labels for different graph nodes. An example of node classification problem in web networks is to predict the categories for web pages, where the web pages are represented as nodes and hyperlinks as edges in a graph \cite{xiong2020memory}. In program analysis for compiler optimization, some data flow analysis tasks have also been formulated as a node classification problem, such as reachability analysis, which predicts if a statement is reachable from a root statement, where statements are represented as nodes in the program graph \cite{cummins2021programl}. In these tasks, how to learn node representations that can best capture the relations or dependencies between nodes is a key problem. A memory mechanism in GNNs that could enhance a GNNs' capability to capture such information may greatly benefit these node classification tasks. \cite{xiong2020memory} has shown by their experiments that incorporating global graph information via the memory in MemGAT and MemGCN significantly improves the GNN's performance in node classification tasks. Similar to the graph classification tasks, these memory models operate over the scope of space and will likely only benefit from forgetting mechanisms if the graph is dynamic.

information about entities and their relations \cite{bordes2015large, moon2019memory}. There are also some works that focus on visual QA systems, where contexts from images are also used for answering the questions \cite{ norcliffe2018learning, khademi2020multimodal}. Memory GNN models can greatly benefit these applications, since they have the capability to store the semantic and structural information in memory and utilize them with knowledge graphs for the QA tasks.

\paragraph{Recommender Systems.}
two different forms to memory to capture both short-term item contextual information and long-term item dependencies. They also incorporate item co-occurrences to model the relationships between closely related items.

\paragraph{Meta-learning on graphs}
has an LSTM memory controller that interacts with an external memory module using a number of read and write heads. It has demonstrated superior performance in meta-learning tasks such as few-shot prediction for image classification \cite{santoro2016meta}. In the meta-learning area, there are also problems that involves relational reasoning, for example, the few-shot human-object interaction recognition task that aims at inferring new interactions between human actions and objects with only a few available instances \cite{ji2021task}. Having a memory-augmented GNN that can learn a general strategy for the structured representations it should place into memory and how it should use these representations for predictions could potentially benefit these tasks.

\paragraph{Reinforcement learning over graphs}
Memory is an important aspect of intelligence and plays an important role in many deep reinforcement learning (RL) models \cite{fortunato2019generalization}. For example, episodic memory has been shown to enable reinforcement learning agents to adapt more quickly and thus improve data efficiency \cite{pritzel2017neural, hansen2018fast}. In graph environments, existing reinforcement learning efforts mainly focus on combinatorial optimization problems on graphs, such as the Maximum Cut \cite{barrett2020exploratory}, community detection \cite{wilder2019end}, and Traveling Salesman problems \cite{khalil2017learning}. Some of the works in this area started to consider memory-augmented GNNs for RL. For instance, the memory-augmented graph RL work \cite{zweig2020neural} we discussed in Section \ref{sec:framework} introduces an external memory adapted to the graph environment to enable a recurrent model with greater representation power for the graph navigation task. However, little progress had been made in augmenting GNNs with memory for these graph reinforcement learning tasks. This would be a promising future direction.

\section{Other Related Work}
\label{sec:otherrelated}

While this review is focused on message-passing GNNs augmented with some form of memory, here we discuss other machine learning approaches that address the same problems in machine learning (e.g. modeling long-range dependencies).

\subsection{Other message-passing neural networks}
Message-passing neural networks (MPNNs) have been the most popular architecture for learning graph-structured data. However due to localized information propagation mechanisms, most MPNNs are limited in their capability to learn long-range relationships in graphs (Section \ref{sec:framework}). Several works have proposed a variety of strategies for capturing long-range dependencies in graphs that do not involve memory. For instance, a recurrent GNN framework called Implicit Graph Neural Networks (IGNN) was introduced in \cite{gu2020implicit}. IGNN defines a fixed-point equation as an implicit layer for aggregation, and uses an iterative solver to find the equilibrium
While this technically allows for an `infinite' number of layers, the effective range of IGNN (i.e. the maximum number of hops to capture dependencies for each node) is bounded by a certain value \cite{liu2022mgnni}. This limitation still hinders their ability to capture long-range dependencies in the graph. \cite{velickovic2019deep} propose an unsupervised node representation learning approached called Deep Graph Infomax (DGI). DGI maximizes mutual information between patch representations and corresponding high-level summaries of graphs to capture the global information content of the entire graph. In \cite{you2019position}, Position-aware Graph Neural Networks (P-GNNs) are proposed to capture the position/location of nodes within a broader context of the graph structure. Specifially, P-GNN first samples sets of anchor nodes, computes the distance of a given target node to each anchor-set, and then learns a non-linear distance-weighted aggregation scheme over the anchor-sets, to capture positions of nodes with respect to the anchor nodes. Although these works have improved GNNs' performance for specific tasks, the major limitations of MPNNs such as over-smoothing and over-squashing may still exist due to the message passing mechanism. It would be beneficial to investigate on how to leverage memory with these MPNNs to further improve their performance and expressive power for different applications.

\subsection{Graph Transformer architectures}
The connection between GNNs and GTs hinges on the definition of permutation invariant operations.
That is, both models are closely related to the DeepSets model which defines objective functions over permutation-invariant inputs, enabling deep learning over sets \cite{zaheer2017deepsets}.

\cite{muller2023attending} investigated how well GTs can recover graph properties. They showed that GTs outperform MPNNs in some long-range problems, demonstrating their potential to prevent over-squashing. Although the GT models have achieved promising results in various graph learning tasks, one major downside of GTs is their scalability to larger graphs. The dense attention mechanism in GTs means that the time complexity scales quadratically with the number of nodes, $\mathcal{O}(N^2)$. Future work should focus on theoretical comparison of GTs and MPNNs on various graph learning tasks, rather than purely experimental comparisons.

\section{Open Challenges and Future Directions}
\label{sec:future}

\subsection{Empirical Evaluations of Memory}
The current evaluations of the memory GNN models in the existing works are mostly done on standard graph learning applications, such as graph/node classification and recommender systems. They use common metrics such as classification accuracy to measure the performance of the models and compare with baseline GNNs. Although they are able to demonstrate the overall performance of the models for the applications with these metrics, they lack a framework with fine-grained measures for evaluating the contribution of memory for these tasks. For instance, it would be useful to have more metrics to measure the long range relationships that existing networks can capture. Although some works have attempted to do this, such as the work in \cite{alon2020bottleneck} that introduces a NeighborsMatch task to evaluate GNNs for long-range problems, there is still a lack of general evaluation frameworks, including benchmark datasets, tasks, and metrics, for the empirical evaluation of the memory GNNs. This would be a valuable future direction.

\subsection{Scalability} The enhanced capabilities of memory-augmented neural networks often come at a high computational cost \cite{stevens2019manna}. Memory GNNs could be even more prone to this issue, since the data they deal with are complex structured data. For instance, the memory mechanism proposed for the MemGCN \cite{xiong2020memory} introduces an extra $O(NM)$ message passings per layer for the GNNs, which allows each of the $N$ nodes in the original graph to interact with the $M$ memory nodes stored in external memory. This extra cost could be very large when scaling to big graphs or when the graph is sparse. In other memory GNN models there is no forgetting mechanism, leading them to maintain memories through their lifetime without discarding any outdated information \cite{strathmann2021persistent}. Depending on the application this can cause issues with complexity and storage costs and severely limits scalability. 
that suffer from complexity issues introduced by the memory augmentation, one promising solution is to have a selective write mechanism that controls what is written to memory.  For example, instead of storing information from all nodes in the MemGCN \cite{xiong2020memory}
, they could consider some form of selection criteria or an attention mechanism to ensure only relevant and useful information is written into the memory. This could improve the efficiency and thus the scalability of the models. 
Furthermore, adding selective forgetting mechanisms to models can 
not only reduce the storage cost of the memory, but it can also help alleviate the influence of outdated information and prevents the model from overfitting to past events.

\section{Conclusion}
We have presented a review of the mechanisms, applications, and limitations of current memory GNN models from the lens of neuroscience and psychology. It is clear that multiple forms of memory exist in these models (Tables \ref{tab:taxonomy1}-\ref{tab:taxonomy2}) to support a variety of structured representation learning problems. In particular, internal memory supports the local persistence of information across time or across a portion of the graph, whereas external memory can provide more global information or contextual knowledge. Future work on memory GNN models may improve both performance and scalability, but these advances depend on the continued development of both theoretical and empirical tools to evaluate the contribution of memory.

\bibliographystyle{IEEEtran}


\end{document}